%% file: main.tex
\documentclass{article}
\usepackage{spconf,amsmath,graphicx}

\usepackage{import}
\usepackage{booktabs}
\usepackage{tablefootnote}
\usepackage{amsfonts}
\usepackage{amssymb}
\usepackage{caption} 
\title{multi-encoder multi-resolution framework for end-to-end\\ speech recognition}
\name{Ruizhi Li$^1$, Xiaofei Wang$^1$, Sri Harish Mallidi$^2$, Takaaki Hori$^3$, Shinji Watanabe$^1$, Hynek Hermansky$^1$}
\address{$^1$The Johns Hopkins University, $^2$Amazon, $^3$Mitsubishi Electric Research Laboratories (MERL)\\
{\small \tt \{ruizhili, xiaofeiwang, shinjiw, hynek\}@jhu.edu, mallidih@amazon.com, thori@merl.com}
}

\begin{document}
\ninept

\maketitle

\import{./}{sec0.abst}
\import{./}{sec1.intro}

\import{./}{sec2.model}

\import{./}{sec3.expt}

\import{./}{sec4.conclusion}

\vfill\pagebreak

\bibliographystyle{IEEEbib}
\bibliography{strings,refs}

\end{document}

%% file: sec0.abst.tex
\begin{abstract}
Attention-based methods and Connectionist Temporal Classification (CTC) network have been promising research directions for end-to-end Automatic Speech Recognition (ASR). 
The joint CTC/Attention model has achieved great success by utilizing both architectures during multi-task training and joint decoding. 
In this work, we present a novel Multi-Encoder Multi-Resolution (MEMR) framework based on the joint CTC/Attention model. 
Two heterogeneous encoders with different architectures, temporal resolutions and separate CTC networks work in parallel to extract complimentary acoustic information. 
A hierarchical attention mechanism is then used to combine the encoder-level information. 
 To demonstrate the effectiveness of the proposed model, experiments are conducted on Wall Street Journal (WSJ) and CHiME-4, resulting in relative Word Error Rate (WER) reduction of $18.0-32.1\%$. 
Moreover, the proposed MEMR model achieves $\textbf{3.6\%}$ WER in the WSJ eval92 test set, which is the best WER reported for an end-to-end system on this benchmark. 

\end{abstract}


%% file: sec1.intro.tex
\section{Introduction}
\label{sec:intro}
Recent advancements in deep neural networks enabled several practical applications of automatic speech recognition (ASR) technology. 
The main paradigm for an ASR system is the so-called hybrid approach, which involves training a DNN to predict context dependent phoneme states (or senones) from the acoustic features. 
During inference the predicted senone distributions are provided as inputs to decoder, which combines with lexicon and language model to estimate the word sequence. 
Despite the impressive accuracy of the hybrid system, it requires hand-crafted pronunciation dictionary based on linguistic assumptions, extra training steps to derive context-dependent phonetic
models, and text preprocessing such as tokenization for languages
without explicit word boundaries. 
Consequently, it is quite difficult for non-experts to develop ASR systems for new applications, especially for new languages.

End-to-End speech recognition approaches are designed to directly output word or character sequences from the input audio signal. 
This model subsumes several disjoint components in the hybrid ASR model (acoustic model, pronunciation model, language model) into a single neural network. 
As a result, all the components of an end-to-end model can be trained jointly to optimize a single objective. 
Two dominant end-to-end architectures for ASR are Connectionist Temporal Classification (CTC) \cite{graves2006connectionist,graves2014towards,miao2015eesen} and attention-based encoder decoder \cite{chan2015listen,chorowski2015attention} models. 
While CTC efficiently addresses sequential problem (speech vectors to word sequence mapping) by avoiding the alignment pre-construction step using dynamic programming, it assumes conditional independence of label sequence given the input. 
Attention model does not assume conditional independence of label sequence resulting in a more flexible model. 
However, attention-based methods encounter difficulty in satisfying the speech-label monotonic property. 
To alleviate this issues, a joint CTC/Attention framework was proposed in\cite{kim2016joint_icassp2017,hori2017advances,watanabe2017hybrid}. 
The joint model was shown to provide the state-of-the-art end-to-end results in several benchmark datasets \cite{watanabe2017hybrid}.

In end-to-end ASR approaches, the encoder acts as an acoustic model providing higher-level features for decoding. 
Bi-directional Long Short-Term Memory (BLSTM) has been widely used due to its ability to model temporal sequences and their long-term dependencies as the encoder architecture; 
Deep convolutional Neural Network (CNN) was introduced to model spectral local correlations and reduce spectral variations in end-to-end framework \cite{hori2017advances,zhang2016very}.
The encoder architecture combining CNN with recurrent layers, was suggested to address the limitation of LSTM. 
While temporal subsampling in RNN and max-pooling in CNN aim to reduce the computational complexity and enhance the robustness, it is likely that subsampling technique results in loss of temporal resolution.

In this work, we propose a Multi-Encoder Multi-Resolution (MEMR) model within the joint CTC/Attention framework. This is strongly motivated by the success of multi-stream paradigm in Hybrid ASR \cite{mallidi2018practical, hermansky2013multistream,mallidi2016novel} mimicking human speech processing cognitive system. Two parallel encoders with heterogeneous structures, RNN-based and CNN-RNN-based, are mutually complementary in characterizing the speech signal. 

Several studies have shown that attention-based model benefits from having multiple attention mechanisms \cite{Hayashi2018,chiu2018state,vaswani2017attention, yang2016hierarchical, hori2017attention,libovicky2017attention}. 
Inspired by the advances in Hierarchical Attention Network (HAN) in document classification~\cite{yang2016hierarchical}, multi-modal video description~\cite{hori2017attention} and machine translation~\cite{libovicky2017attention}, we adapt HAN into our MEMR model.  
The encoder that carries the most discriminate information for the prediction can dynamically receive a stronger weight.  
Each encoder is associated with a CTC network to guide the frame-wise alignment process for individual encoder. 

This paper is organized as follows: 
section 2 explains the joint CTC/Attention model. 
The description of the proposed MEMR framework is in section 3. 
Experiments with results and several analyses are presented in section 4. 
Finally, in section 5 the conclusion is derived.

%% file: sec2.model.tex
\section{Joint CTC/Attention Mechanism}
\label{ssec:ctcatt}
In this section, we review the joint CTC/attention architecture, which takes advantage of both CTC and attention-based end-to-end ASR approaches during training and decoding.

\subsection{Connectionist Temporal Classification (CTC)}
\label{sssec:ctc}

Following Bayes decision theory, CTC enforces a monotonic mapping from a $T$-length speech feature sequence, $X=\{\textbf{x}_{t}\in \mathbb{R}^{D}|t = 1,2,...,T\}$, to an $L$-length letter sequence, $C=\{c_{l}\in \mathcal{U}|l = 1,2,...,L\}$. Here $\textbf{x}_{t}$ is a $D$-dimensional acoustic vector at frame $t$, and $c_{l}$ is at position $l$ a letter from $\mathcal{U}$, a set of distinct letters. 

The CTC network introduces a many-to-one function from frame-wise latent variable sequences, $Z=\{z_{t}\in \mathcal{U} \bigcup \text{blank} |t=1,2,...,T\}$, to letter predictions of shorter lengths.
Note that the additional ``blank'' symbol is used to handle the merging of repeating letters. With several conditional independence assumptions, the posterior distribution, $p(C|X)$, is represented as follows:
\begin{equation}
\label{f:ctcloss}
p(C|X)\approx \sum_{Z}\prod_{t} p(z_{t}|X) \triangleq p_{ctc}(C|X),
\end{equation}
where $p(z_{t}|X)$ is a frame-wise posterior distribution, and we also define the CTC objective function $p_{ctc}(C|X)$.
CTC preserves the benefits that it avoids the HMM/GMM construction step and preparation of pronunciation dictionary.



\subsection{Attention-based Encoder-Decoder}
\label{sssec:att}

As one of the most commonly used sequence modeling techniques, the attention-based framework selectively encodes an audio sequence of variable length into a fixed dimension vector representation, which is then consumed by the decoder to produce a distribution over the outputs.  
We can directly estimate the posterior distribution $p(C|X)$ using the chain rule:
\begin{equation}
p(C|X)=\coprod_{l=1}^{L}p(c_{l}|c_{1},...,c_{l-1}, X) \triangleq p_{att}(C|X),
\end{equation}
where $p_{att}(C|X)$ is defined as the attention-based objective function. 
Typically, a BLSTM-based encoder transforms the speech vectors $X$ into frame-wise hidden vector $\textbf{h}_{t}$
If the encoder subsamples the input by a factor $s$, there will be $T/s$ time steps in $H=\{\textbf{h}_{1},..., \textbf{h}_{T/s}\}$. 
The letter-wise context vector $\textbf{r}_{l}$ is formed as a weighted summation of frame-wise hidden vectors $H$ using content-based attention mechanism. 

In comparison to CTC, not requiring conditional independence assumptions is one of the advantages of using the attention-based model. 
However, the attention is too flexible to satisfy monotonic alignment constraint in speech recognition tasks.

\subsection{Joint CTC/Attention}
\label{sssec:ctcatt}

The joint CTC/Attention architecture benefits from both CTC and attention-based models since the attention-based encoder-decoder is trained together with CTC within the Multi-Task Learning (MTL) framework. 
The encoder is shared across CTC and attention-based encoders.  
And the objective function to be maximized is a logarithmic linear combination of the CTC and attention objectives, i.e., $p_{ctc}(C|X)$ and $p_{att}^{\dagger}(C|X)$:
\begin{equation} 
\label{f:mtl}
\mathcal{L}_{MTL}=\lambda\log p_{ctc}(C|X)+(1-\lambda)\log p_{att}^{\dagger}(C|X),
\end{equation}
where $\lambda$ is a tunable scalar satisfying $0\leq\lambda\leq 1$. $p_{att}^{\dagger}(C|X)$ is an approximated letter-wise objective where the probability of a prediction is conditioned on previous true labels. 


During inference, the joint CTC/Attention model performs a label-synchronous beam search.
The most probable letter sequence $\hat{C}$ given the speech input $X$ is computed according to
\begin{align}
\label{f:jointdec}
    \hat{C}=\arg\max_{C\in \mathcal{U}^{*}} &\{\lambda \log p_{ctc}(C|X)+(1-\lambda)\log p_{att}(C|X) \nonumber \\
    &+\gamma \log p_{lm}(C)\} 
\end{align}
where external RNN-LM probability $\log p_{lm}(C)$ is added with a scaling factor $\gamma$. 

\section{Proposed MEMR framwork}
\label{ssec:memodel}

\begin{figure}[htb]
  \centering 
  \centerline{\includegraphics[width=8.5cm]{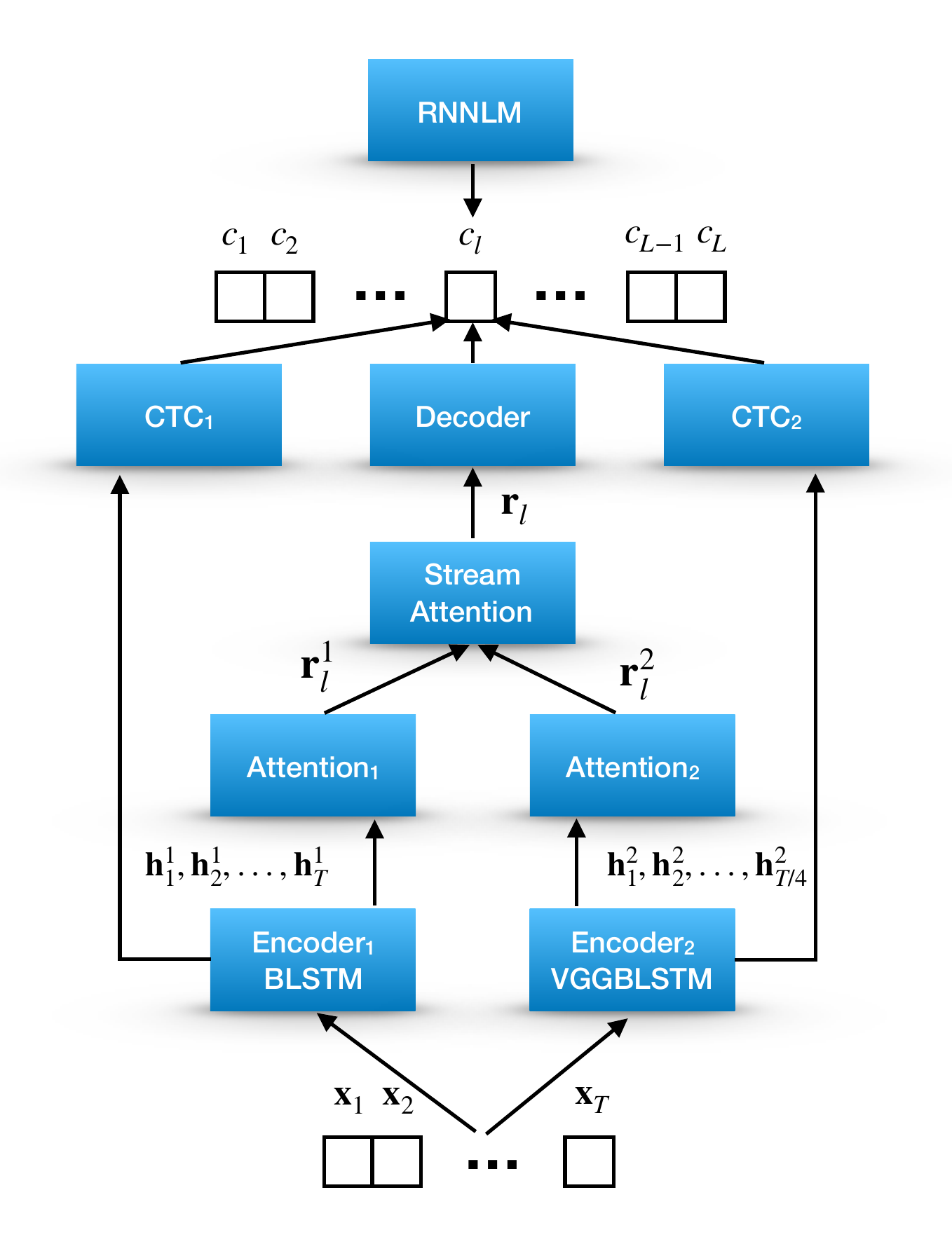}}
\caption{The Multi-Encoder Multi-Resolution Architecture.}
  \label{fig:me}
\end{figure}

The overall architecture is shown in Fig. \ref{fig:me}. 
Two types of encoders with different temporal resolutions are presented in parallel to capture acoustic information in various ways, followed by an attention fusion mechanism together with per-encoder CTC. 
An external RNN-LM is also involved during the inference step. 
We will describe the details of each component in the following sections. 

\subsection{Multi-Encoder with Multi-Resolution}
\label{sssec:encarch}
We propose a Multi-Encoder Multi-Resolution (MEMR) architecture that has two encoders, RNN-based and CNN-RNN-based.
Both encoders take the same input features in parallel operating on different temporal resolutions, aiming to capture complimentary information in the speech. 

The RNN-based encoder is designed to model temporal sequences with their long-range dependencies. 
In MEMR, the BLSTM encoder has only BLSTM layers that extract the frame-wise hidden vector $\textbf{h}_t^1$ without subsampling in any layer:
\begin{equation}
\label{f:blstm}
\textbf{h}_{t}^{1}=\textrm{Encoder}_{1}(X)\triangleq \textrm{BLSTM}_{t}(X)
\end{equation}
where the BLSTM decoder is labeled as index $1$. 

The combination of CNN and RNN allows the convolutional feature extractor applied on the input to reveal local correlations in both time and frequency dimensions. 
The RNN block on top of CNN makes it easier to learn temporal structure from the CNN output, to avoid modeling direct speech features with more underlying variations.
The pooling layer is essential in CNN to reduce the spatial size of the representation to control over-fitting.
In MEMR, we use the initial layers of the VGG net architecture~\cite{simonyan2014very} followed by BLSTM layers as VGGBLSTM decoder labeled as index 2:
\begin{equation}
\label{f:vggblstm}
\textbf{h}_{t}^{2}=\textrm{Encoder}_{2}(X)\triangleq \textrm{VGGBLSTM}_{t}(X).
\end{equation}
The configuration of convolutional layers in VGGBLSTM encoder is the same as in \cite{hori2017advances}. 

\subsection{Hierarchical Attention}
\label{sssec:hieratt}

Since the encoders in MEMR describe the speech signal differently by catching acoustic knowledge in their own ways, encoder-level fusion is suitable to boost the network's ability to retrieve the relevant information.  
We adapt Hierarchical Attention Network (HAN) in \cite{yang2016hierarchical} for information fusion. 
The decoder with HAN is trained to selectively attend to appropriate encoder, based on the context of each prediction in the sentence as well as the higher-level acoustic features from both encoders, to achieve a better prediction. 

The letter-wise context vectors, $\textbf{r}_l^1$ and $\textbf{r}_l^2$, from individual encoders are computed as follows: 
\begin{equation}
\label{f:cv1}
\textbf{r}_{l}^{1}={\sum}_{t=1}^{T}a_{lt}^1\textbf{h}_{t}^{1},\  \textbf{r}_{l}^{2}={\sum}_{t=1}^{T/4}a_{lt}^2\textbf{h}_{t}^{2},
\end{equation}
where the attention weights are obtained using a scontent-based attention mechanism. Note that since $\textrm{Encoder}_2$ performs downsampling by 4, the summation is till $T/4$ in Eq. (\ref{f:cv1}).

The fusion context vector $\textbf{r}_l$ is obtained as a convex combination of $\textbf{r}_l^1$ and $\textbf{r}_l^2$ as illustrated in the following:
\begin{equation}
\label{f:han}
\textbf{r}_{l}=\beta_{l1}\textbf{r}_{l}^{1} + \beta_{l2}\textbf{r}_{l}^{2},
\end{equation}
\begin{equation}
\label{f:l2att}
{\beta}_{li}=\textrm{ContentAttention}(\textbf{q}_{l-1}, \textbf{r}_l^i), i=1,2.
\end{equation}
The stream-level attention weights $\beta_{l1}$ and $\beta_{l2}$ are estimated according to the previous decoder state, $\textbf{q}_{l-1}$, and context vectors, $\textbf{r}_l^1$ and $\textbf{r}_l^2$, from individual encoders as described in Eq. (\ref{f:l2att}).
The fusion context vector is then fed into the decoder to predict the next letter. 

\subsection{Per-encoder CTC}
\label{sssec:perctc}
In the CTC/Attention model with a single encoder, the CTC objective serves as an auxiliary task to speed up the procedure of realizing monotonic alignment and providing a sequence-level objective. 
In the MEMR framework, we introduce per-encoder CTC where a separate CTC mechanism is active for each encoder stream during training and decoding.
Sharing one set of CTC among encoders is a soft constraint that limits the potential of diverse encoders to reveal complimentary information.
In the case that both encoders are with different temporal resolutions and network architectures, per-encoder CTC can further align speech with labels in a monotonic order and customize the sequence modeling of individual streams.

During training and decoding steps, we follow Eq. (\ref{f:mtl}) and (\ref{f:jointdec}) with a change of the CTC objective $\log p_{ctc}(C|X)$ in the following way:
\begin{equation}
\log p_{ctc}(C|X)=\frac{1}{2}\lambda(\log p_{ctc_1}(C|X)+\log p_{ctc_2}(C|X)),
\end{equation}
where joint CTC loss is the average of per-encoder CTCs. 

%% file: sec3.expt.tex
\section{experiments}
\label{sec:expt}

\subsection{Experimental Setup}
\label{ssec:exptsetup}
We demonstrate our proposed MEMR model using two datasets: WSJ1~\cite{wsj1} (81 hours) and CHiME-4~\cite{vincent20164th} (18 hours). 
In WSJ1, we used the standard configuration: ``si284'' for training, ``dev93'' for validation, and ``eval92'' for test. 
The CHiME-4 dataset is a noisy speech corpus recorded or simulated using a tablet equipped with 6 microphones in four noisy environments: a cafe, a street junction, public transport, and a pedestrian area. 
For training, we used both ``tr05\_real'' and ``tr05\_simu'' with additional WSJ1 corpora to support end-to-end training. 
``dt05\_multi\_isolated\_1ch\_track'' is used for validation. 
We evaluated the real recordings with 1, 2, 6-channel in the evaluation set. The BEAMFORMIT method was applied to multi-channel evaluation.
In all experiments, 80-dimensional mel-scale filterbank coefficients with additional 3-dimensional pitch features served as the input features.
\begin{table}[ht]
  \begin{center}
   	\caption{Comparison among single-encoder end-to-end models with BLSTM or VGGBSLTM as the encoder, the MEMR model and prior end-to-end models. (WER: WSJ1, CHiME-4)}
	\begin{tabular}{lcc}
	  \toprule
	  \toprule
	   &\multicolumn{1}{c}{CHiME-4} & \multicolumn{1}{c}{WSJ1}  \\
	   Model & et05\_real\_1ch & eval92 \\
	  \midrule
       {\it BLSTM (Single-Encoder)} &  &\\
      CTC   & 62.7 &36.4 \\
      ATT   & 50.2 &20.8 \\
      CTC+ATT  & 29.2 &4.6 \\
      \midrule
      {\it VGGBLSTM (Single-Encoder) } &  \\
      CTC   & 50.6 &19.1 \\
      ATT   & 42.2 &17.2\\
      CTC+ATT   & 29.6 &5.6 \\
      \midrule
      {\it BLSTM+VGGBLSTM (MEMR) } &  \\
      CTC   & 49.1 &15.2 \\
      ATT   & 44.3 &18.9 \\
      CTC(shared)+ATT   & 26.8 &4.4 \\
      CTC(shared)+ATT+HAN   & 26.9 &4.3 \\
      CTC(per-enc)+ATT  & 26.6 &4.1 \\
      CTC(per-enc)+ATT+HAN   & \textbf{26.4} &\textbf{3.6} \\
      \midrule
      {\it Previous Studies } &  \\
      RNN-CTC \cite{graves2014towards} & - & 8.2\\
      Eesen \cite{miao2015eesen} & - & 7.4 \\
      Temporal LS + Cov. \cite{chorowski2016towards} & - & 6.7 \\
      E2E+regularization\cite{zhou2017improved} & -& 6.3 \\
      Scatt+pre-emp\cite{zeghidour2018end} & -& 5.7 \\
      Joint e2e+look-ahead LM\cite{hori2018end} & -& 5.1 \\
      RCNN+BLSTM+CLDNN \cite{wang2017residual}& -&4.3\\
      EE-LF-MMI \cite{hadian2018end} &-&4.1 \\
      \bottomrule
      \bottomrule
	\end{tabular}
	\label{tab:memr}
  \end{center}
\end{table}

The $\textrm{Encoder}_1$ contains four BLSTM layers, in which each layer has 320 cells in both directions followed by a 320-unit linear projection layer. The $\textrm{Encoder}_2$ combines the convolution layers with RNN-based network that has the same architecture as $\textrm{Encoder}_1$. A content-based attention mechanism with 320 attention units is used in encoder-level and frame-level attention mechanisms. The decoder is a one-layer unidirectional LSTM with 300 cells. We use 50 distinct labels including 26 English letters and other special tokens, i.e., punctuations and sos/eos.

We incorporated the look-ahead word-level RNN-LM~\cite{hori2018end} of 1-layer LSTM with 1000 cells and 65K vocabulary, that is, 65K-dimensional output in Softmax layer. 
In addition to the original speech transcription, the WSJ text data with 37M words from 1.6M sentences was supplied as training data.
RNN-LM was trained separately using Stochastic Gradient Descent (SGD) with learning rate $=0.5$ for 60 epochs. 

The MEMR model is implemented using Pytorch backend on ESPnet.
Training procedure is operated using the AdaDelta algorithm with gradient clipping on single GPUs, ``GTX 1080ti''. 
The mini-batch size is set to be 15.
We also apply a unigram label smoothing technique to avoid over-confidence predictions.
The beam width is set to 30 for WSJ1 and 20 for CHiME-4 in decoding. 
For model jointly trained with CTC and attention objectives, $\lambda=0.2$ is used for training, and $\lambda=0.3$ for decoding.
RNN-LM scaling factor $\gamma$ is $1.0$  for all experiments with the exception of using $\gamma=0.1$ in decoding attention-only models.

\subsection{Results}
\label{ssec:results}

The overall experimental results on WSJ1 and CHiME-4 are shown in Table \ref{tab:memr}.
Compared to joint CTC/Attetion single-encoder models, the proposed MEMR model with per-encoder CTC and HAN achieves relative improvements of $9.6\%$ ($28.4\%\rightarrow 26.4\%$) in CHiME-4 and 21.7\% in WSJ1 ($4.6\%\rightarrow 3.6\%$) in terms of WER. 
We compare the MEMR model with other end-to-end approaches, and it outperforms all of the systems from previous studies. 
We design experiments with fixed encoder-level attention ${\beta}_{l1}={\beta}_{l2}=0.5$. And the MEMR model with HAN outperforms the ones without parameterized stream attention.
Moreover, per-encoder CTC constantly enhances the performance with or without HAN. Specially in WSJ1, the model shows notable decrease ($4.3\%\rightarrow 3.6\%$) in WER with per-encoder CTC. 
Our results further confirms the effectiveness of joint CTC/Attention architecture in comparison to models with either CTC or attention network. 

\begin{table}[!htbp]
  \begin{center}
   	\caption{Comparison between the MEMR model and VGGBSLTM single-encoder model with similar network size. (WER: WSJ1, CHiME-4)}
	\begin{tabular}{lcc}
	  \toprule
	  \toprule
	  & Single-Encoder & Proposed Model\\
	  Data & (21.9M) & (21.3M) \\
      \midrule
      {\it CHiME-4} & \\
      et05\_real\_1ch & 32.2 & \textbf{26.4 (18.0\%)} \\
      et05\_real\_2ch & 26.8 & \textbf{21.9 (18.3\%)} \\
      et05\_real\_6ch & 21.7 & \textbf{17.2 (20.8\%)} \\
      \midrule
	  {\it WSJ1}  & \\
      eval92 & 5.3 & \textbf{3.6 (32.1\%)} \\
      \bottomrule
      \bottomrule
	\end{tabular}
	\label{tab:cmp1stream}
  \end{center}
\end{table}
For fair comparison, we increase the number of BLSTM layers from 4 to 8 in $\textrm{Encoder}_2$ to train a single-encoder model. 
In Table \ref{tab:cmp1stream}, the MEMR system outperforms the single-encoder model by a significant margin with similar amount of parameters, $21.9$M v.s. $21.3$M.
In CHiME-4, we evaluate the model using real test data from 1, 2, 6-channel resulting in an average of $\textbf{19\%}$ relative improvement from all three setups.
In WSJ1, we reach $\textbf{3.6\%}$ WER in eval92 in our MEMR framework with relatively $\textbf{32.1\%}$ improvement.


\begin{table}[!htbp]
  \begin{center}
   	\caption{Effect of Multi-Resolution Configuration $(s_1,s_2)$, where  $s_1$ and $s_2$ are the subsampling factors for $\textrm{Encoder}_1$ and $\textrm{Encoder}_2$, respectively. (WER: WSJ1, CHiME-4)}
	\begin{tabular}{lccc}
	  \toprule
	  \toprule
	  Data & (4,4) & (2,4) & (1,4) \\
	  \midrule
      {\it CHiME-4} & & & \\	  
      et05\_real\_1ch & 29.1& 27.0 &\textbf{26.4}\\
      \midrule
      {\it WSJ1}  & & \\
      eval92 &  4.5& 4.2&\textbf{3.6}\\
      \bottomrule
      \bottomrule
	\end{tabular}
    \label{tab:ss}
  \end{center}
\end{table}
The results in Table \ref{tab:ss} shows the contribution of multiple resolution.  
The WER goes up when increasing subsampling factor $s_1$ closer to $s_2=4$ in both datasets. In other words, the fusion works better when two encoders are more heterogeneous which supports our hypothesis. 
As shown in Table \ref{tab:att}, We analyze the average stream-level attention weight for $\textrm{Encoder}_2$ when we gradually decrease the number of LSTM layers while keeping $\textrm{Encoder}_1$ with the original configuration. It aims to show that HAN is able to attend to the appropriate encoder seeking for the right knowledge. As suggested in the table, more attention goes to $\textrm{Encoder}_1$ from $\textrm{Encoder}_2$ as we intentionally make $\textrm{Encoder}_2$ weaker.
\begin{table}[!htbp]
  \begin{center}
   	\caption{Analysis of Hierarchical Attention mechanism when when fixing $\textrm{Encoder}_1$ and changing the number of LSTM layers in $\textrm{Encoder}_2$. (WER: CHiME-4)}
    \label{tab:table5}
	\begin{tabular}{c|cc}
	  \toprule
	  \toprule
      \# LSTM Layers &\multicolumn{1}{c}{Average Stream Attention} &   \\
	  in VGGBLSTM & for VGGBLSTM & WER \% \\
	  \midrule
	   0& 0.27 & 30.6\\
       1& 0.52 & 29.8\\
       2 & 0.75 & 28.9\\
	   3 & 0.82 & 27.8\\
       4 & 0.81 & 26.4\\
      \bottomrule
      \bottomrule
	\end{tabular}
	\label{tab:att}
  \end{center}
\end{table}

%% file: sec4.conclusion.tex
\section{conclusion}
\label{sec:conclusion}

In this work, we present our MEMR framework to build an end-to-end ASR system. 
Higher-level frame-wise acoustic features are carried out from RNN-based and CNN-RNN-based encoders with subsampling only in convolutional layers. 
Stream fusion selectively attends to each encoder via a content-based attention.  
We also investigated that assigning a CTC network to individual encoder further enhance the heterogeneous configuration of encoders. 
The MEMR model outperforms various single-encoder models, reaching the state-of-the-art performance on WSJ among end-to-end systems. 